\documentclass[3p,times]{elsarticle}
\usepackage{ecrc}
\usepackage{graphicx}
\usepackage{diagbox}
\volume{}
\usepackage{lineno,hyperref}
\modulolinenumbers[5]
\firstpage{1}
\journalname{Neurocomputing}
\runauth{}
\jid{procs}
\jnltitlelogo{Procedia Computer Science}
\CopyrightLine{}{}
\usepackage{amssymb}
\usepackage{float}

\usepackage{graphicx} % 需使用包
\usepackage{algorithm}  
\usepackage{algpseudocode}  
\usepackage{amsmath}  
\begin{document}

\begin{frontmatter}

\dochead{}

\title{Neural Architecture Search based on the Cartesian Genetic Programming}

%% use optional labels to link authors explicitly to addresses:
%% \author[label1,label2]{<author name>}
%% \address[label1]{<address>}
%% \address[label2]{<address>}

\author[1,2]{Xuan Wu} 
\author[1,2]{Xiuyi Zhang}  
\author[1,2]{Linhan Jia}   
\author[3]{Liang Chen}  
\author[4]{Yanchun Liang}
\author[1,2,3]{You Zhou\corref{aa}} 
\author[1,2]{Chunguo Wu\corref{aa}}

\address[1]{Key Laboratory of Symbolic Computation and Knowledge Engineering of Ministry of Education,}
\address[2]{College of Computer Science and Technology, Jilin University, Changchun, 130012, China}
\address[3]{College of Software, Jilin University, Changchun, 130012, China}
\address[4]{School of Computer Science, Zhuhai College of Science and Technology, Zhuhai, 519041, China}
	\cortext[aa]{\{wucg, zyou\}@jlu.edu.cn}
	%%* Corresponding Author: {wucg, zyou}@jlu.edu.cn

\begin{abstract}
	
Neural architecture search (NAS) is a hot topic in the field of automated machine learning and outperforms humans in designing neural architectures on quite a few machine learning tasks. Motivated by the natural representation form of neural networks by Cartesian Genetic Programming (CGP), we propose an evolutionary approach of NAS based on CGP, called CGPNAS, to solve sentence classification task. To evolve the architectures under the framework of CGP, the operations such as convolution are identified as the types of function nodes of CGP, and the evolutionary operations are designed based on Evolutionary Strategy. The experimental results show that the searched architectures are comparable with the performance of human-designed architectures. We verify the ability of domain transfer of our evolved architectures and the transfer experimental results show that the accuracy deterioration is lower than 2-5\%. Finally, the ablation study identifies the Attention function as the single key function node and the linear transformations along could keep the accuracy similar with the full evolved architectures, which is worthy of investigation in the future.
\end{abstract}

%% keywords here, in the form: keyword \sep keyword

%% PACS codes here, in the form: \PACS code \sep code

%% MSC codes here, in the form: \MSC code \sep code
%% or \MSC[2008] code \sep code (2000 is the default)

\end{frontmatter}

%%
%% Start line numbering here if you want
%%
% \linenumbers

%% main text
\section{INTRODUCTION}
\label{sec1}
As a core technique in modern data-driven artificial intelligence, Deep Neural Networks (DNNs) have surpassed the achievement of former methods in many typical problems and have made excellent solutions to questions in interdisciplinary research. However, the architecture designing of DNNs is limited by the existing knowledge of designers, which makes it hard to find the global best architecture for a given task. Hence, much attention has been paid to the Neural Architecture Search (NAS) to relieve the burden of researchers from architecture design for DNNs and to best explore the architecture searching space \cite{ref1,ref2}. There are many methods proposed to search architecture, among which Reinforcement Learning (RL) and Evolutionary Algorithm (EA) are the most popular. 

Zoph et al. (2017) \cite{ref3} firstly use the policy gradient algorithm, a RL approach, as the Recurrent Neural network (RNN) controller to produce new architectures of Convolution Neural Network (CNN). Subsequently, Zoph et al. (2018) \cite{ref4} use the RL with proximal policy optimization as the RNN controller. Baker et al. (2017) \cite{ref5} use Q-learning with the $\epsilon$-greedy exploration strategy to sequentially search for neural architectures. To relieve expensive calculations on GPUs, several speed-up methods and efficient solutions are proposed based on the RNN controller. Pham et al. (2018) \cite{ref6} propose  Efficient Neural Architecture Search (ENAS), in which the controller searches for the best subgraph within a larger graph in the first stage and shares parameters between subgraphs in the second stage. Compared with the original work in \cite{ref3}, ENAS accelerates the efficiency of NAS up to a thousand times.

As another popular method, NAS based on EA has a history of more than 30 years. Gruau (1993) \cite{ref7} proposes Cellular Encoding (CE), which is a grammatical inference process to search neural networks with Genetic Programming (GP). Yao and Liu (1997) \cite{ref8} propose Evolutionary Programming Network (EPNet), which evolves the network architecture and connection weights with Evolutionary Programming. To evolve neurons of a network, Stanley and Miikkulainen (2002) \cite{ref9} propose NeuroEvolution of Augmenting Topologies (NEAT), which encodes the neurons into Node genes and Connection genes and uses Genetic Algorithm (GA) to update Node genes and Connection genes. 

With the emergence of Automated Machine Learning, many NAS methods based on EA are proposed in recent years. Xie et al. (2017) \cite{ref10} propose GeNet based on GA to choose CNNs, where CNN is divided into different stages with pooling operation as the boundary, and all convolution operations in the same stage have the same convolution kernel and channel number. Suganuma et al. (2019) \cite{ref11} use Cartesian Genetic Programming (CGP), a graph form of GP, to encode the CNN architectures (CGP-CNN). CGP-CNN adopts highly functional Block as the node functions, for example, ConvBlock including convolution, batch normalization, and ReLU. Bi et al. (2019) \cite{ref12} propose Feature Learning GP (FLGP) to evolve convolution operators for feature learning on image classifications. Sun et al. (2019) \cite{ref13} use PSO to search Flexible Convolutional Auto Encoders (FCAE) with chain structure. To search image classifier, Real et al. (2019) \cite{ref14} modify the tournament selection evolution by introducing an age property to favor the younger genotypes (named Aging Evolution or regularized evolution), which keeps as many young individuals as possible. 

Most of the NAS methods are proposed to solve Computer Vision (CV) problems \cite{ref15} and focus on evolving CNN architectures. Nowadays, researchers make efforts to enable NAS to solve problems in the field of Natural Language Processing (NLP). Since Transformer \cite{ref16} has become the state-of-the-art model in NLP, David et al. (2019) \cite{ref17} use Transformer as initial, design a new searching space for NLP problems, and search for the best candidate Transformers, named Evolved Transformer. Ramakanth et al. (2020) \cite{ref18} propose Flexible and Expressive Neural Architecture Search (FENAS), dividing the search process into two stages similar to ENAS\cite{ref6}. The results show FENAS can reproduce Long Short-Term Memory (LSTM) and Gated Recurrent Unit (GRU) structures.

Sentence classification task is a classical and fundamental task in the field of NLP. Motivated by the effectiveness of Transformer for NLP problems and the natural representation of DNN with CGP, this paper proposes a CGP encoding-based NAS (CGPNAS) method to deal with sentence classification task.

The remaining parts are organized as follows: Section \ref{sec2} introduces the related work briefly; Section \ref{sec3} proposes the CGPNAS; Section \ref{sec4} presents the experimental results to evaluate the performance of CGPNAS; and finally, Section \ref{sec5} presents the conclusion.

\section{RELATED WORK}
\label{sec2}
In this Section, we first introduce how EA applied to NAS in Part \ref{sec2.1}. Next, we briefly review the research on sentence classification task in Part \ref{sec2.2}. Finally, we introduce CGP method in Part \ref{sec2.3}, which is the encoding method of this paper .

\subsection{Neural architecture search based on Evolutionary Algorithms}
\label{sec2.1}
NASs based on EA mainly focus on the following two aspects: encoding method and genetic operator. The encoding method is to convert the phenotype into the genotype of a given DNN. The genetic operator is to produce new genotypes in each iteration. Except encoding method and genetic operator, there are also a small number of studies on survival selection strategy and parental selection strategy \cite{ref14,ref19}.
In EA, there are two classic kinds of genetic operators: Crossovers and Mutations. Crossovers combine the genotype of two or more parents to get one or more offspring genotypes. Mutations change the genotype of a parent to get a new genotype. To produce new genotypes, different NASs use one or both two kinds of genetic operators. CGP-CNN \cite{ref11} use mutation as genetic operator only. NEAT \cite{ref9}, GeNet \cite{ref10}, FLGP \cite{ref12}, AmoebaNet-A \cite{ref14} and DCNN designer \cite{ref20} use both crossover and mutation as genetic operators.

There are two types of encoding methods: direct and indirect. As a widely used method, the direct encoding methods explicitly specify neural architecture information with genotypes. In NEAT \cite{ref9}, the genotype is composed of Node genes and Connection genes. Node genes store the node type, indicating input (or sensor) node, output node or hidden node. Connection genes store the numbers of in-nodes and out-nodes, and the weights, states (Enabled or Disenabled) and innovation numbers of the connections. Because FCAE is used to evolve a chain architecture without explicit topological connection information, Sun et al. \cite{ref13} only encode node type and its parameter information into genotype. The indirect encoding methods specify only a generating rule of genotypes. CE \cite{ref7} is a classical indirect encoding method. The entire neural network evolves from a single ancestor cell where the evolutionary DNA is stored in a tree structure. The tree structure defines the method of cell division, generating the final network topology with cell development.

\subsection{Sentence classification task}
\label{sec2.2}
Sentence classification is a classical and fundamental task in NLP. Traditional classification methods often use human-designed features, which could learn only the shallow representation of sentences. With the development of deep learning, CNN, RNN and Attention \cite{ref21} are widely used in sentence classification tasks. Hochreiter and Schmidhuber (1997) \cite{ref22} propose Long Short-term Memory (LSTM) as a special RNN for long-term dependencies learning, which relieves the gradient disappearance effectively in the process of back propagation by gating mechanism. After that, there is great success in dealing with NLP problems with LSTM. Kim (2014) \cite{ref23} apply a simple CNN to sentence classification task, and achieve excellent results on multiple benchmarks. Compared with traditional machine learning methods, Kim’s method is good at capturing location features of sentences. Vaswani et al. \cite{ref16} (2017) propose the Transformer architecture based on Attention mechanisms, which could be widely used in NLP tasks. BERT \cite{ref24} is a pre-trained architecture, characterized by Masked Language Model and Next Sentence Prediction, which could create the state-of-the-art performance for a wide range of tasks by finetuning just the output layer. 

Since different methods have their own advantages, many scholars combine multiple methods to achieve competitive results than a single method. Lai et al. \cite{ref25} (2015) propose Recurrent Convolutional Neural Networks, combining the bidirectional RNN and max-pooling layer in CNN. Liu and Guo \cite{ref26} (2019) propose AC-BiLSTM, combining the Attention mechanism, convolutional layer and bidirectional LSTM. Zhang et al. \cite{ref27} (2019) propose 3W-CNN, combining deep learning methods and traditional feature-based methods. Zhang et al. design a confidence function to divide the outputs of CNN into 2 parts with strong and weak confidence, respectively. The CNN classification outputs with weak confidence will be reclassified by NB-SVM proposed in \cite{ref28}.

\subsection{Cartesian genetic programming}
\label{sec2.3}
As a graph form of GP, CGP is initially proposed to optimize digital circuits \cite{ref29}, and hence, each intermediate node has two inputs. Subsequently, CGP is applied to many problems, such as image processing and molecular docking \cite{ref30,ref31,ref32,ref33}. 

As shown in Fig. 1, CGP is represented by a directed graph with n input nodes, m output nodes. Except input nodes and output nodes, CGP has $r\ast c$ intermediate nodes, also known as function nodes, where r and c denote rows size and Columns size, respectively. CGP could set the number of inputs for each function node, for example, Ref. \cite{ref29} sets 2 as the number of inputs and the number of outputs is usually set 1. In addition, CGP forbids the links in the same grid columns, and usually sets a max stride of connection between columns, called “levels-back”, which can increase or reduce the size of searching space. Borrowing the words in genetics, some function nodes, e.g., Node $a_{r,c}$, is not be used as input for subsequent nodes, called inactive nodes \cite{ref11}.
\begin{figure}[H]
	\centering
	\label{fig1}
	\includegraphics[scale=0.2]{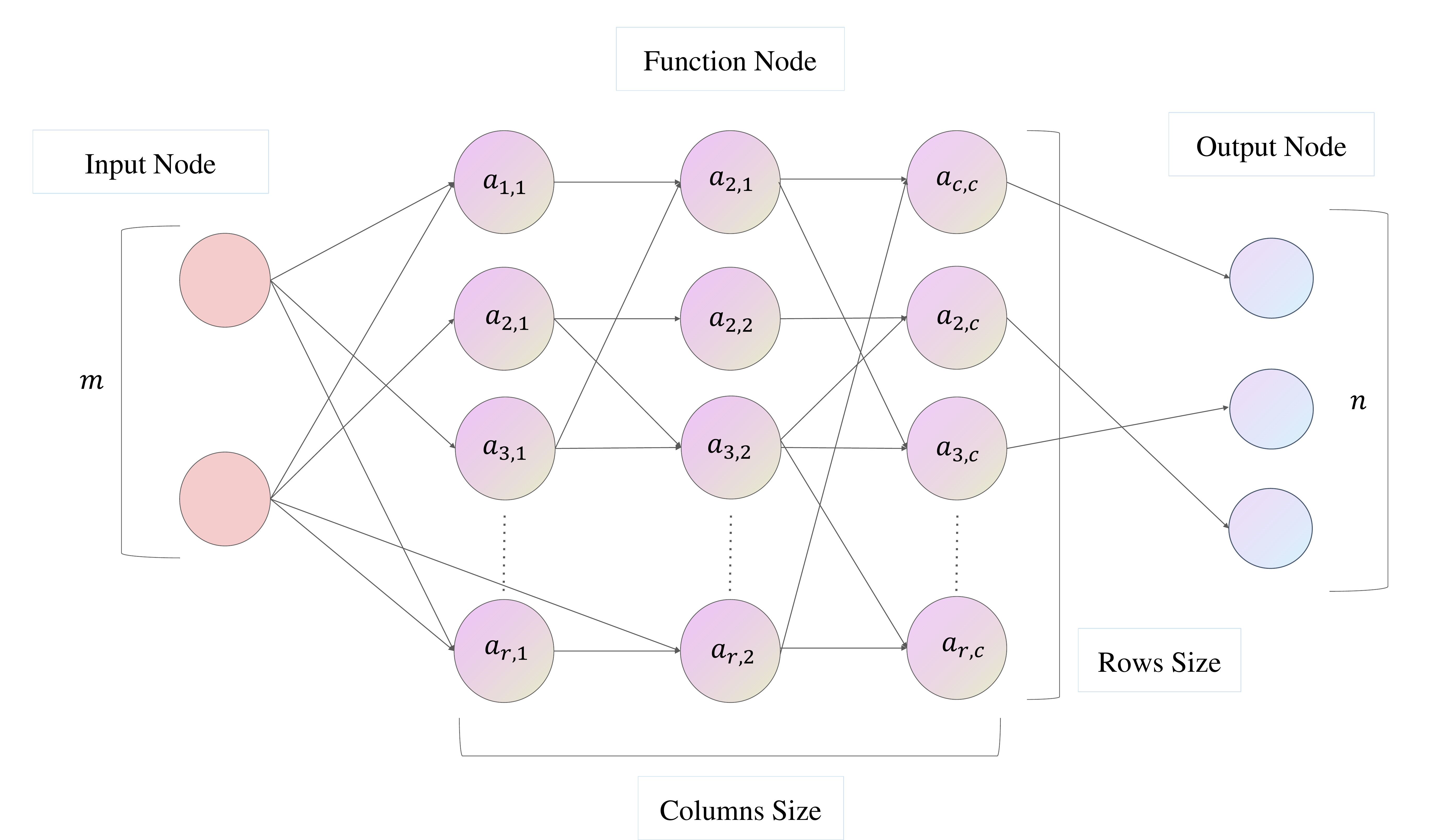}
	\caption{Illustration of Cartesian Genetic Programming.}
\end{figure}

\section{OUR METHOD}
\label{sec3}
Motivated by the natural representation form of neural networks by CGP, we propose a novel NAS method based on CGP, named CGPNAS, to deal with sentence classification task. In Part \ref{sec3.1}, we introduce CGP coding method applied to NAS. In Part \ref{sec3.2}, we present the function nodes used in this paper. In Part \ref{sec3.3}, we design an evolution method for CGPNAS.

\subsection{CGP coding method}
\label{sec3.1}
For NAS problem, CGP uses a two-dimensional grid as the phenotype of neural networks, as shown in Fig. 1, which is a natural presentation of neural networks due to the topological similarity between CGP and neural networks. The links represent the data flow and the function nodes represent basic operations of the neural networks, such as Convolution, Attention and so on. 

The encoding structure of CGP is a triplet shown at the bottom in Fig. 2-a, indicating the function name and the two numbers of input nodes. An illustrating genotype, with 10 function nodes, is shown above the encoding structure in Fig. 2-a. For the genotype, each gene corresponds to a node in Fig. 2-b, which is the intermediate phenotype with both inactive and active links in dashed and solid arrows, respectively. In addition, Node 6 and Node 7 are both inactivate nodes. Fig. 2-c is the final phenotype with only active links in solid arrows and can be used as a DNN to solve problems. 
\begin{figure}[H]
\label{fig2}
	\centering
	\includegraphics[scale=0.25]{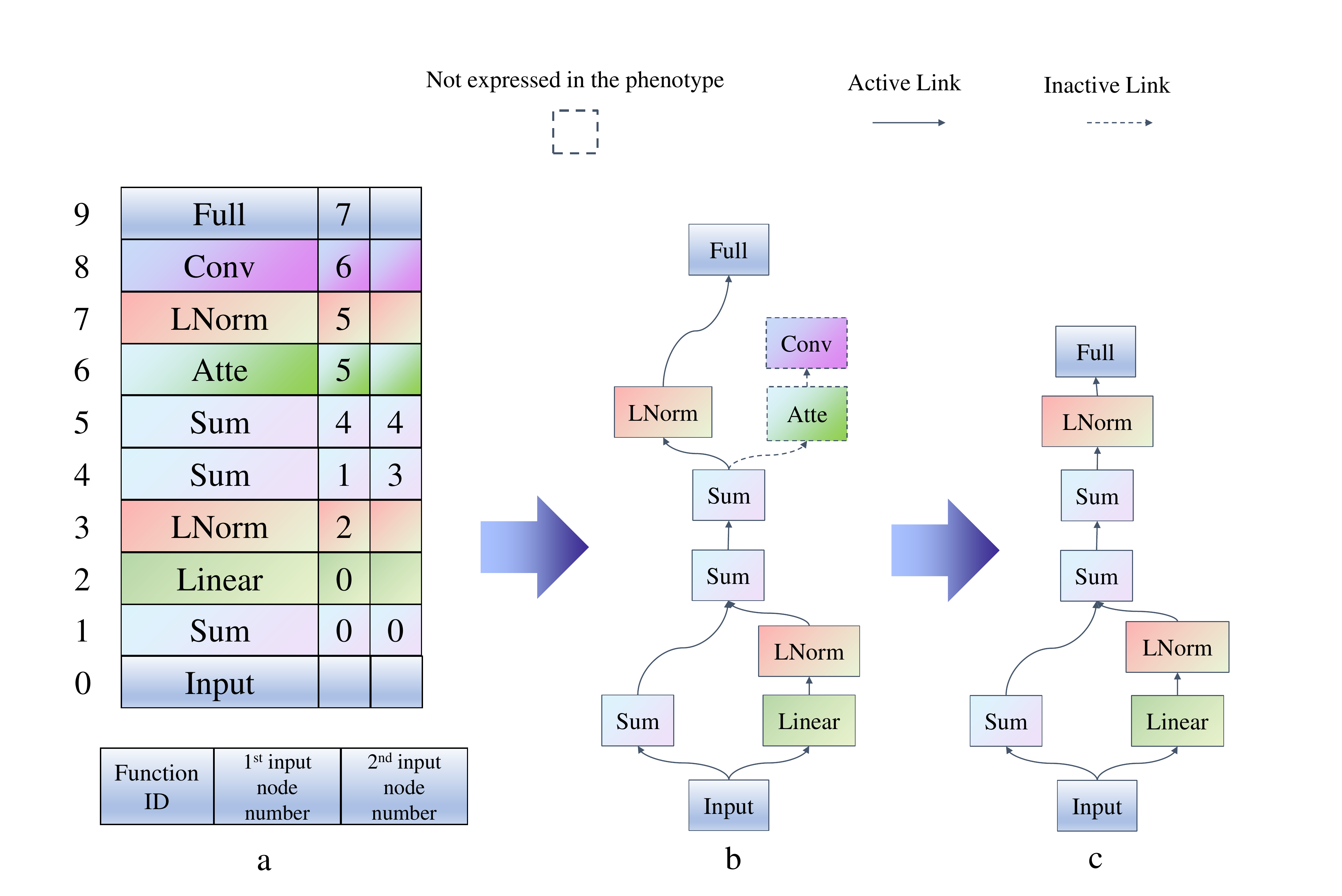}
	\caption{Illustration of Cartesian Genetic Programming.}
\end{figure}

\subsection{Function Node Design}
\label{sec3.2}
The set of function nodes is important for the evolved neural architectures. Hence, for the task of sentence classification, we design the set of function nodes as follows, denoted as S:
$$
S=\{Conv,\ Atte,\ Linear,\ Sum,\ ReLU,\ LNorm,\ GLU\}
$$
where the enumerated symbols mean the operations of Convolution, Attention, Linear, Sum, ReLU, Layer Normalization \cite{ref35} and Gated Linear Units (GLU) \cite{ref36}. The function node types, the number of input nodes, parameter name, candidate parameter values, input and outputs dimensions are shown sequentially in Table 1, from the fifth to the sixth column, where $b$, $l$, $d$, and $d^\prime$ denote batch size, max sentence length, input dimension of word vectors and output dimension of word vectors.  It is worth emphasizing that when the node has two inputs, we use $d_1$ and $d_2$  to represent the word vector dimensions of the two inputs respectively. 
\begin{table}[H]
\label{table1}
	\centering
	\caption{THE TYPES OF FUNCTION NODES AND THEIR CANDIDATE PARAMETER VALUES.}
	\begin{tabular}{l|l|l|l|l|l}
		\hline
		Node Type & \# Input Nodes & Para. Name & Para. Value  & Input Dim. & Output Dim. \\\hline
		Conv.     & 1              & Channel    & \{16, 32\}   & $b\times l\times d$ &   $b\times l\times d'$          \\
		&                & Kernel     & \{1, 3, 5\}  &            &             \\\hline
		Atte.     & 1              & Head       & \{4, 8, 16\} &        $b\times l\times d$    &        $b\times l\times d$     \\\hline
		Linear.    & 1              & Channel    & \{32, 128\}  &       $b\times l\times d$     &       $b\times l\times d'$      \\\hline
		Sum.       & 2              & -          & -            &        $b\times l\times d_1$    &      $b\times l\times d'$       \\
		 &              &           &             &        $b\times l\times d_2$    &             \\\hline
		ReLU.      & 1              & -          & -            &      $b\times l\times d$      &        $b\times l\times d$     \\\hline
		LNorm.    & 1              & -          & -            &      $b\times l\times d$      &          $b\times l\times d$   \\\hline
		GLU.       & 1              & -          & -            &     $b\times l\times d$       &            $b\times l\times d'$ \\\hline

	\end{tabular}
\end{table}

For the task of sentence classification, one-dimensional convolutions and multi-head attention are used. The Linear node represents a linear transformation. The function of Sum node is to merge two branches. When two branches with different dimensions of word vectors are going to be merged, the smaller word vector would be filled with 0 at its end to force it into the same size as the larger one. Although Sum node has two input nodes formally, it is allowed to receive the same input two times from a single precursor Sum node, such as Node 1 and Node 5, shown in Fig. 2-a. The Layer Normalization is proposed by Ba et, al. \cite{ref35} for RNN, which is normalized in the channels and features of samples. The Linear node represents a linear transformation. GLU node is a variant of Convolution with gate-controlled outputs.
\subsection{Evolution strategy design}
\label{sec3.3}
CGP usually uses the $1+\lambda$ Evolutionary Strategy (ES) to update and select the population, meaning that one parental individual and $\lambda$ offspring individuals compete to survive into the next generation. Through mutation operation and adaptive selection, the population evolves towards the optimal goal. According to \cite{ref11}, there are two kinds of mutations in 1+$\lambda$ ES, named forced mutation and neutral mutation, respectively. The forced mutation works on all parental nodes to generate offspring, and the neutral mutation works only on inactive parental nodes to contribute potentially new nodes for the next generation. Both forced mutation and neutral mutation are point mutations, which means that the function and connection of nodes randomly change to valid values according to the mutation rate. To enhance exploration and overcome the local optimal traps, we double the initial mutation rate for the late 25\% generation.

The algorithm is described as follows. Firstly, the $\lambda$ offspring individuals are produced by the current parental individual through the forced mutation. If all fitness of the $\lambda$ offspring individuals are worse than their parental individual, the inactive nodes of the parental individual are mutated by neutral mutation, and the $\lambda$ offspring individuals are eliminated. Otherwise, the offspring individual with the highest fitness is selected as the parental individual of the next generation. The pseudocode is presented as follows:
  \begin{algorithm}[htb]  
	\caption{ Evolution Strategy}    
	\begin{algorithmic}[1]  
		\State Create a parent randomly
		\State Evaluate the fitness of parent
		\While{ generation $\textless$ Max\_generation}
		\State Double the mutation rate for late 25\% generation
		\State $\lambda$ offspring are produced by forced mutation.
		\State Evaluate the fitness of $\lambda$ offspring individuals
		\If{the $\lambda$ offspring individuals are all worse than the parent}
		\State Mutate the inactive nodes of parent with neutral mutation
		 \Else{
		  offspring with the best fitness become the new parent for the next iteration}
		 \EndIf
		 
	\EndWhile 
	\State End
	\end{algorithmic}  
\end{algorithm} 

The accuracy of sentence classification task corresponding to each architecture is taken as the individual fitness. The neutral mutation acts on inactive nodes, it does not change the parental fitness, so we do not need to evaluate the altered parent by the neutral mutation.

\section{EXPERIMENT}
\label{sec4}
In this Section, we first introduce datasets, hyperparameter and experimental setting details in Part \ref{sec4.1} and \ref{sec4.2}. Next, we compare the searched architecture obtained by CGPNAS and CGPNAS(GloVe) with the classical architecture in Part \ref{sec4.3}. GloVe \cite{ref37} is an embedding method, that allows neural networks not to learn the correlation between words from scratch, so GloVe can improve the performance of the network. And then we verify the transfer ability of the searched architecture on different datasets in Part \ref{sec4.4}. Finally, we implement ablation testing to analyze the impact of function nodes on the searched architecture in Part \ref{sec4.5}.

\subsection{Datasets}
\label{sec4.1}
The following datasets are used in our experiments, shown in Table 2. There are 3 datasets labeled with positive and negative, including SST2 \cite{ref38} (Binary labeled version of Stanford sentiment treebank), MR \cite{ref39} (a large movie review dataset extracted from Rotten Tomatoes web) and IMDB \cite{ref40}. Samples of SST5 \cite{ref38} (Stanford Sentiment Treebank) are labeled with 5 levels, i.e., very positive, positive, neutral, negative and very negative. Samples of AG\_news \cite{ref41}, extracted by ComeToMyHead website, are labeled with 4 kinds of tags, i.e., World, Sports, Business and Sci/Tech.
\begin{table}[H]
\label{table2}
	\centering
	\caption{PROPERTIES OF THE EXPERIMENTAL DATASETS.}
	\begin{tabular}{l|l|l|l|l|l}
		\hline
		Dataset               & SST2 & SST5 & MR & IMDB & Ag\_news \\\hline
		Label levels          & 5    & 2    & 2  & 2    & 4        \\\hline
		max sentence length   & 50   & 50   & 50 & 400  & 50       \\\hline
		word vector dimension & \multicolumn{5}{c}{300} \\\hline   
	\end{tabular}
\end{table}
\subsection{Hyperparameter and experiment details}
\label{sec4.2}
The CGP parameters are shown in Table 3. Initially, we set the CGP grid by $5\times20$ and use a relatively large number of columns size to generate deep architectures. To leverage searching space complexity and models’ generalization ability, Levels-back is set to 3. The lower and upper bounds of numbers of active nodes are 10 and 60, respectively. To enhance the exploration ability, the offspring size is set to 4. 

\begin{table}[H]
\label{table3}
	\centering
	\caption{EXPERIMENTAL PARAMETERS.}
	\begin{tabular}{l | l}
		\hline
		Parameters&Values         \\
		\hline
		Input nodes number&1      \\
		Input nodes number&1      \\
		Rows Size r&5              \\
		Columns Size c&20                       \\
		 Levels-back&3           \\
		Activate nodes number&$[10, 60]$           \\
		Mutation rate&$\{0.1, 0.2, 0.4\}$ \\
		Offspring Size $\lambda$&4   \\
		Max\_generation&1000        \\\hline

	\end{tabular}
\end{table}

To enhance exploration, we set the initial mutation rate of the early 75\% generations as 0.1, and double it into 0.2 for the late 25\% generations. However, the mutation rate of SUM function nodes should be larger than that of other functional nodes to decrease the probability of single-chain architectures. Hence, we set the mutation rate of SUM function nodes as 0.2 and 0.4 by trials, respectively, in the early and late generations.

Taking the time consumption into account, we try to use the small values for the max sentence length and the word vector dimension as shown in Table 2. However, due to the average sentence length of IMDB is 8 times larger than the other dataset, the max sentence length is set as 400. For all experimental datasets, the word vector dimension is set uniformly as 300.

In Parts \ref{sec4.3},\ref{sec4.4} and \ref{sec4.5} we train CGPNAS and CGPNAS(GloVe) with Adam Optimizer for 50 epochs and the learning rate is 0.01. In Part \ref{sec4.3}, the classic architectures in comparison include TextCNN \cite{ref23}, Transformer \cite{ref16}, BERT \cite{ref24}, Evolved Transformer \cite{ref17}, AC-BiLSTM \cite{ref26}, 3W-CNN \cite{ref27} and FENAS \cite{ref18}. 

In this paragraph, we introduce the training details of the comparison algorithm in Part \ref{sec4.3}. Similar to CGPNAS, we also train TextCNN, Transformer, BERT and Evolved Transformer with Adam Optimizer for 50 epochs and the learning rate is 0.01. In addition, We train a 6 layers Transformer encoder \cite{ref17} and the number of attention heads is set to 6. We follow the official guide from \cite{ref42} to finetune the BERT-Base-Uncased model \cite{ref24} for downstream tasks. We use the searched network from \cite{ref17}, training a 6 layer Evolved Transformer encoder with a linear layer to perform classification task at last.

\subsection{Comparation with other algorithms}
\label{sec4.3}
To present the performance of CGPNAS and CGPNAS(GloVe) on different datasets and perform statistical tests, we execute CGPNAS and CGPNAS(GloVe) 10 times on each dataset, respectively. As an example, one of the searched architectures on IMBD dataset is shown in Table 4. 

As shown in Table 5, with the help of GloVe, CGPNAS(GloVe) knows the correlation between words in the initial stage and improves the accuracy by 2-5\% on different datasets, compared with CGPNAS. Hence, the performance of CGPNAS is similar to TextCNN and the performance of CGPNAS(GloVe) is similar to Transformer and Evolved Transformer. As the human-designed architectures, BERT and AC-BiLSTM get the best accuracy on 2 and 3 datasets, respectively. It can be said that for the sentence classification task, even if the existing NASs can reach the human-designed level, they are still difficult to outperform the best human-designed methods.

\begin{table}[H]
\label{table4}
	\centering
	\caption{DIMENSION CHANGE OF THE SEARCHED NEURAL NETWORK.}
	\begin{tabular}{ll|ll}
		\hline
		
		\multicolumn{2}{c}{Input}                       &
		 \multicolumn{2}{c}{$8\times400\times300$}  \\\hline
		Sum                          &$8\times400\times300$  & Linear (Channel: 128) & $8\times400\times128$ \\\hline
		Conv (Channel: 32 Kernel: 1) & $8\times400\times32$  & LNorm                 & $8\times400\times128$  \\\hline
	\multicolumn{2}{c}{Sum}                      &\multicolumn{2}{c}{$8\times400\times128$}  \\\hline
		\multicolumn{2}{c}{Sum}                      &\multicolumn{2}{c}{$8\times400\times128$}  \\\hline
		\multicolumn{2}{c}{Atte (Head: 16)}                       &\multicolumn{2}{c}{$8\times400\times128$}  \\\hline
		\multicolumn{2}{c}{LNorm}                    &\multicolumn{2}{c}{$8\times400\times128$}  \\\hline
		\multicolumn{2}{c}{Atte (Head: 4)}                       &\multicolumn{2}{c}{$8\times400\times128$}  \\\hline
		\multicolumn{2}{c}{Conv (Channel: 32 Kernel: 3)}                       &\multicolumn{2}{c}{$8\times400\times32$}  \\\hline
		\multicolumn{2}{c}{Conv (Channel: 16 Kernel: 5)} &  \multicolumn{2}{c}{$8\times400\times16$}  \\\hline
\end{tabular}
\end{table}
\begin{table}[H]
\label{table5}
	\centering
	\caption{COMPARISON OF DIFFERENT ALGORITHMS. (“*” RESULTS FROM THE ORIGINAL PAPERS. THE CELLS HIGHLIGHTED IN BOLD INDICATE THE BEST ACCURACY)}
	\begin{tabular}{l|l|l|l|l|l}
		\hline
		\diagbox{Architecture}{Dataset}                   & SST2  & SST5  & MR    & IMDB  & Ag\_news \\\hline
		 
		TextCNN (2014)             & 0.812 & 0.372 & 0.713 & 0.84  & 0.817    \\\hline
		Transformer (2017)         & 0.855 & 0.365 & 0.746 & 0.863 & 0.853    \\\hline
		BERT(2019)                       & $\mathbf{0.915}$ & 0.423 & 0.821 & 0.912 & $\mathbf{0.892}$    \\\hline
		Evolved Transformer (2019) & 0.769 & 0.385 & 0.717 & 0.873 & 0.812    \\\hline
		AC-BiLSTM* (2019)          & 0.883 & $\mathbf{0.489}$ & $\mathbf{0.832}$ & $\mathbf{0.918}$ & -        \\\hline
		3W-CNN* (2019)&	-	&-	&0.823&	-&	-\\\hline
		FENAS* (2020)&	0.866&	-	&-&	-&	-\\\hline
		CGPNAS&	$0.733\pm0.027$&	$0.362\pm0.006$&	$0.704\pm0.015$&	$0.844\pm0.012$	&$0.843\pm0.017$\\\hline
		CGPNAS
		(GloVe)	&$0.788\pm0.013$&	$0.413\pm0.013$&	$0.744\pm0.015$&	$0.864\pm0.011$&	$0.864\pm0.018$
		     \\\hline
	\end{tabular}
\end{table}
\subsection{Transfer ability study}
\label{sec4.4}
To verify the transfer ability of the searched architecture, we transfer all the architectures searched on one dataset to the other datasets. The results are shown in Table 6.

We can see that the architectures searched on Ag\_news still perform better on the target dataset; the mean of accuracy improves 1\% on target datasets SST2 and MR, reduces by 1\% on target dataset SST5 and reduces by 3\% on target dataset  IMDB. But the architectures searched on SST2, SST5, MR and IMDB perform slightly poorly on target datasets, the mean of accuracy is reduced by 2-5\%. In particular, on the target dataset Ag\_news, the mean of most accuracy is reduced by 7-8\%, especially 15\% of architectures searched on SST5.  The results show that the architecture searched by CGPNAS has transfer ability and can be applied to most target datasets, but the accuracy of some target datasets has decreased significantly.

\begin{table}[H]
\label{table6}
	\centering
	\caption{TRANSFER TESTING OF CGPNAS.}
	\begin{tabular}{l|l|l|l|l|l}
		\hline
	\diagbox{Origin}{Target}&
		SST2&	SST5&	MR&	IMDB&	Ag\_news\\\hline
	SST2&	$0.733\pm0.027$&	$0.324\pm0.022$&	$0.661\pm0.020$& $0.814\pm0.009$&	$0.762\pm0.057$\\\hline
	SST5&	$0.673\pm0.015$&	$0.362\pm0.006$&	$0.654\pm0.017$&	$0.797\pm0.018$&	$0.689\pm0.021$\\\hline
	MR	&$0.706\pm0.018$&	$0.324\pm0.013$&	$0.704\pm0.015$&	$0.813\pm0.008$&$0.769\pm0.046$\\\hline
	IMDB&	$0.689\pm0.044$	&$0.341\pm0.011$	&$0.674\pm0.014$	&$0.844\pm0.012$&	$0.776\pm0.044$\\\hline
	Ag\_news&	$0.742\pm0.026$&	$0.351\pm0.020$&	$0.711\pm0.018$&	$0.819\pm0.011$&	$0.843\pm0.017$
	
		\\\hline
	\end{tabular}
\end{table}

\subsection{Ablation study}
\label{sec4.5}
To investigate the key component that has a remarkable contribution to the performance, the ablation testing is presented in this part. For this purpose, we reduce the diversity of functions in the set of function nodes and create three new sets of function nodes. The first one is denoted as $S \backslash \{Conv\}$, which removes Convolution in the set of function node. The second one is denoted as $S \backslash \{Atte\}$, which removes Attention in set of function node. The third one is denoted as $S \backslash \{Conv,Atte\}$, which removes both Convolution and Attention in the set of function node. We execute CGPNAS 10 times on each set of function nodes, respectively. Schematic architectures of ablation testing are shown in Fig. 3.

It can be seen from Table 7. that even if the Convolution is removed, the accuracy improves 0.6\% on IMDB and drops only by 1-2\% on the rest datasets. However, if the Attention is removed, the average accuracy drops by 1.1\% on Ag\_news but by 4-6\% on the other datasets. The experimental results show that the Attention function node is vital for the searched architecture. While it is also noted that even if all Convolution and Attention nodes are both removed, the accuracy drops by 4-5\%. However, the accuracies of the evolved architectures, excluding Convolution and Attention nodes, are higher than those only Attention excluded on SST2, SST5 and MR. It can be known that the architecture shown in Fig. 3-c performs mainly the linear transformation from its input, but it still achieves better accuracy than $S\{Atte\}$. The detailed mechanism is worthy of investigation in the future. 

\begin{table}[H]
\label{table7}
	\centering
	\caption{ABLATION TESTING OF CGPNAS.}
	\begin{tabular}{l|l|l|l|l|l}
		\hline
		
		\diagbox{Search Space}{Dataset}&	SST2&	SST5&	MR&	IMDB&	Ag\_news\\\hline
		$S \backslash \{Conv\}$&	$0.717\pm0.018$&	$0.348\pm0.003$&	$0.678\pm0.029$&	$0.850\pm0.005$&	$0.838\pm0.006$\\\hline
	$S\backslash\{Atte\}$&	$0.678\pm0.022$&	$0.319\pm0.003$&	$0.647\pm0.008$&	$0.798\pm0.016$&	$0.832\pm0.007$\\\hline
$S\backslash\{Conv,Atte\}$&$	0.690\pm0.019$&	$0.325\pm0.003$&	$0.663\pm0.027$&	$0.795\pm0.009$&	$0.823\pm0.006$\\\hline
S&$	0.733\pm0.027$&	$0.362\pm0.006$&$	0.704\pm0.015$&	$0.844\pm0.012$&	$0.843\pm0.017$\\\hline

	\end{tabular}
\end{table}

\begin{figure}[H]
\label{fig3}
	\centering
	\includegraphics[scale=0.25]{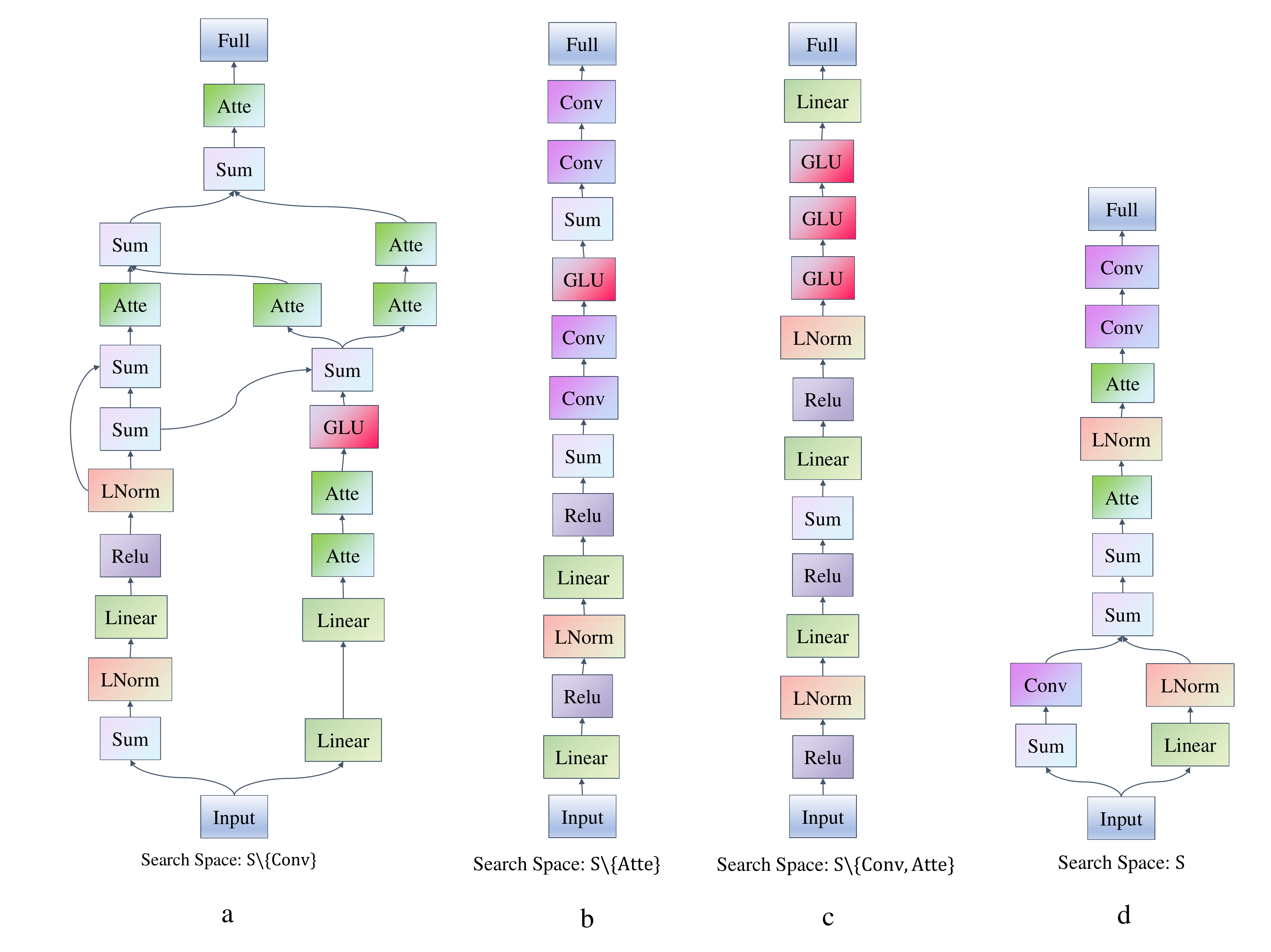}
	\caption{Schematic architectures of ablation testing.}
\end{figure}
\section{CONCLUSION}
\label{sec5}
CGP is a natural representation of neural networks and can evolve the structure and parameters of neural architectures at the same time. For this reason, we propose CGPNAS which can reach the state-of-the-art of human-designed architectures for sentence classification tasks. The transfer study proves that the evolved architectures have transfer ability and can be applied to different target domains. According to the ablation testing, the attention mechanism is very important for CGPNAS, which also proves the reason why the attention mechanism is widely used in NLP.

NAS is still worthy of in-depth study on NLP and subsequent work can increase the diversity of function, such as adding LSTM in the set of function node. To give the design specification of the neural network, a large number of experiments can be carried out to give which combinations are more likely to appear in the network. In addition, the basic mathematical operations can be considered as function nodes to expand the representation ability of the evolved architectures.

\section*{Acknowledgement}

This work is supported by the National Natural Science Foundation of China (61876069, 61972174 and 61972175), the Jilin Natural Science Foundation (20200201163JC), and the Science and Technology Planning Project of Guangdong Province (2020A0505100018), the Guangdong Key-Project for Applied Fundamental Research (2018KZDXM076).
%% The Appendices part is started with the command \appendix;
%% appendix sections are then done as normal sections
%% \appendix

%% \section{}
%% \label{}

%% References
%%
%% Following citation commands can be used in the body text:
%% Usage of \cite is as follows:
%%   \cite{key}         ==>>  [#]
%%   \cite[chap. 2]{key} ==>> [#, chap. 2]
%%

%% References with BibTeX database:

%% Authors are advised to use a BibTeX database file for their reference list.
%% The provided style file elsarticle-num.bst formats references in the required Procedia style

%% For references without a BibTeX database:

% \begin{thebibliography}{00}

%% \bibitem must have the following form:
%%   \bibitem{key}...
%%

% \bibitem{}

% \end{thebibliography}


\begin{thebibliography}{99}
	\bibitem{ref1} H. Gu, G. Fu, J. Li, and J. Zhu, “Auto-ReID+: Searching for a multi-branch ConvNet for person re-identification,” Neurocomputing, vol. 435, pp. 53–66, May 2021, doi: 10.1016/j.neucom.2020.12.105.
	\bibitem{ref2} D. Tian, J. Deng, G. Vinod, T. V. Santhosh, and H. Tawfik, “A constraint-based genetic algorithm for optimizing neural network architectures for detection of loss of coolant accidents of nuclear power plants,” Neurocomputing, vol. 322, pp. 102–119, Dec. 2018, doi: 10.1016/j.neucom.2018.09.014.
	\bibitem{ref3} B. Zoph and Q. V. Le, “Neural Architecture Search with Reinforcement Learning,” in 5th International Conference on Learning Representations, ICLR 2017, Toulon, France, April 24-26, 2017, Conference Track Proceedings, 2017. [Online]. Available: https://openreview.net/forum?id=r1Ue8Hcxg
	\bibitem{ref4}B. Zoph, V. Vasudevan, J. Shlens, and Q. V. Le, “Learning Transferable Architectures for Scalable Image Recognition,” in 2018 IEEE/CVF Conference on Computer Vision and Pattern Recognition, Jun. 2018, pp. 8697–8710. doi: 10.1109/CVPR.2018.00907.
	\bibitem{ref5}B. Baker, O. Gupta, N. Naik, and R. Raskar, “Designing Neural Network Architectures using Reinforcement Learning,” in 5th International Conference on Learning Representations, ICLR 2017, Toulon, France, April 24-26, 2017, Conference Track Proceedings, 2017. [Online]. Available: https://openreview.net/forum?id=S1c2cvqee
	\bibitem{ref6}H. Pham, M. Y. Guan, B. Zoph, and Q. V. Le, “Efficient Neural Architecture Search via parameter Sharing,” in 35th International Conference on Machine Learning, ICML 2018, 2018, vol. 9, pp. 6522–6531.
	\bibitem{ref7}F. Gruau, “Cellular encoding as a graph grammar,” in IEE Colloquium on Grammatical Inference: Theory, Applications and Alternatives, 1993, p. 17/1-1710.
	\bibitem{ref8}X. Yao and Y. Liu, “A new evolutionary system for evolving artificial neural networks,” IEEE Trans. Neural Netw., vol. 8, no. 3, pp. 694–713, 1997, doi: 10.1109/72.572107.
	\bibitem{ref9}K. O. Stanley and R. Miikkulainen, “Evolving Neural Networks through Augmenting Topologies,” Evol. Comput., vol. 10, no. 2, Art. no. 2, Jun. 2002, doi: 10.1162/106365602320169811.
	\bibitem{ref10}L. Xie and A. Yuille, “Genetic CNN,” in 2017 IEEE International Conference on Computer Vision (ICCV), Oct. 2017, pp. 1388–1397. doi: 10.1109/ICCV.2017.154.
	\bibitem{ref11}M. Suganuma, M. Kobayashi, S. Shirakawa, and T. Nagao, “Evolution of Deep Convolutional Neural Networks Using Cartesian Genetic Programming,” Evol. Comput., pp. 1–23, Mar. 2019, doi: 10.1162/evco\_a\_00253.
	\bibitem{ref12}Y. Bi, B. Xue, and M. Zhang, “An Evolutionary Deep Learning Approach Using Genetic Programming with Convolution Operators for Image Classification,” in 2019 IEEE Congress on Evolutionary Computation (CEC), 2019, pp. 3197–3204. doi: 10.1109/CEC.2019.8790151.
	\bibitem{ref13}Y. Sun, B. Xue, M. Zhang, and G. G. Yen, “A Particle Swarm Optimization-Based Flexible Convolutional Autoencoder for Image Classification,” IEEE Trans. Neural Netw. Learn. Syst., vol. 30, no. 8, pp. 2295–2309, Aug. 2019, doi: 10.1109/TNNLS.2018.2881143.
	\bibitem{ref14}E. Real, A. Aggarwal, Y. Huang, and Q. V. Le, “Regularized Evolution for Image Classifier Architecture Search,” Proc. AAAI Conf. Artif. Intell., vol. 33, pp. 4780–4789, Jul. 2019, doi: 10.1609/aaai.v33i01.33014780.
	\bibitem{ref15}X. He, K. Zhao, and X. Chu, “AutoML: A survey of the state-of-the-art,” Knowl.-Based Syst., vol. 212, p. 106622, Jan. 2021, doi: 10.1016/j.knosys.2020.106622.
	\bibitem{ref16}A. Vaswani et al., “Attention is All You Need,” in Proceedings of the 31st International Conference on Neural Information Processing Systems, Red Hook, NY, USA, 2017, pp. 6000–6010.
	\bibitem{ref17}D. So, Q. Le, and C. Liang, “The Evolved Transformer,” in Proceedings of the 36th International Conference on Machine Learning, Jun. 2019, vol. 97, pp. 5877–5886. [Online]. Available: http://proceedings.mlr.press/v97/so19a.html
	\bibitem{ref18}R. Pasunuru and M. Bansal, “FENAS: Flexible and Expressive Neural Architecture Search,” in Findings of the Association for Computational Linguistics: EMNLP 2020, Online, Nov. 2020, pp. 2869–2876. doi: 10.18653/v1/2020.findings-emnlp.258.
	\bibitem{ref19}H. Liu, K. Simonyan, O. Vinyals, C. Fernando, and K. Kavukcuoglu, “Hierarchical Representations for Efficient Architecture Search,” in 6th International Conference on Learning Representations, ICLR 2018, 2018. [Online]. Available: https://openreview.net/forum?id=BJQRKzbA-
	\bibitem{ref20}B. Ma, X. Li, Y. Xia, and Y. Zhang, “Autonomous deep learning: A genetic DCNN designer for image classification,” Neurocomputing, vol. 379, pp. 152–161, Feb. 2020, doi: 10.1016/j.neucom.2019.10.007.
	\bibitem{ref21}D. Bahdanau, K. Cho, and Y. Bengio, “Neural Machine Translation by Jointly Learning to Align and Translate,” in 3rd International Conference on Learning Representations, ICLR 2015, San Diego, CA, USA, May 7-9, 2015, Conference Track Proceedings, 2015. [Online]. Available: http://arxiv.org/abs/1409.0473
	\bibitem{ref22}S. Hochreiter and J. Schmidhuber, “Long short-term memory,” Neural Comput., vol. 9, no. 8, pp. 1735–1780, 1997.
	\bibitem{ref23}Y. Kim, “Convolutional Neural Networks for Sentence Classification,” in Proceedings of the 2014 Conference on Empirical Methods in Natural Language Processing (EMNLP), Doha, Qatar, 2014, pp. 1746–1751. doi: 10.3115/v1/D14-1181.
	\bibitem{ref24}J. Devlin, M.-W. Chang, K. Lee, and K. Toutanova, “BERT: Pre-training of Deep Bidirectional Transformers for Language Understanding,” in Proceedings of the 2019 Conference of the North American Chapter of the Association for Computational Linguistics: Human Language Technologies, Volume 1 (Long and Short Papers), Minneapolis, Minnesota, 2019, pp. 4171–4186. doi: 10.18653/v1/N19-1423.
	\bibitem{ref25}S. Lai, L. Xu, K. Liu, and J. Zhao, “Recurrent Convolutional Neural Networks for Text Classification,” in AAAI, 2015, pp. 2267–2273. [Online]. Available: http://www.aaai.org/ocs/index.php/AAAI/AAAI15/paper/view/9745
	\bibitem{ref26}G. Liu and J. Guo, “Bidirectional LSTM with attention mechanism and convolutional layer for text classification,” Neurocomputing, vol. 337, pp. 325–338, Apr. 2019, doi: 10.1016/j.neucom.2019.01.078.
	\bibitem{ref27}Y. Zhang, Z. Zhang, D. Miao, and J. Wang, “Three-way enhanced convolutional neural networks for sentence-level sentiment classification,” Inf. Sci., vol. 477, pp. 55–64, Mar. 2019, doi: 10.1016/j.ins.2018.10.030.
	\bibitem{ref28}S. Wang, “An improved PTAS approximation algorithm for k-means clustering problem,” in 2012 2nd International Conference on Uncertainty Reasoning and Knowledge Engineering, Jalarta, Indonesia, Aug. 2012, pp. 90–94. doi: 10.1109/URKE.2012.6319592.
	\bibitem{ref29}J. MILLER, “Designing electronic circuits using evolutionary algorithms. arithmetic circuits: A case study,” Genet. Algorithms Evol. Strateg. Engineeing Comput. Sci., 1998.
	\bibitem{ref30}J. Rothermich and J. Miller, “Studying the Emergence of Multicellularity with Cartesian Genetic Programming,” in Late Breaking Papers at the Genetic and Evolutionary Computation Conference (GECCO-2002, 2002, pp. 397–403.
	\bibitem{ref31}T. Arslan, “Evolvable Components—From Theory to Hardware Implementations,” Genet. Program. Evolvable Mach., vol. 6, no. 4, Art. no. 4, Dec. 2005, doi: 10.1007/s10710-005-3718-x.
	\bibitem{ref32}A. B. Garmendia-Doval, S. D. Morley, and S. Juhos, “Post Docking Filtering Using Cartesian Genetic Programming,” in Artificial Evolution, Berlin, Heidelberg, 2004, pp. 189–200. doi: 10.1007/978-3-540-24621-3\_16.
	\bibitem{ref33}A. B. Garmendia-Doval, J. F. Miller, and S. D. Morley, “Cartesian Genetic Programming and the Post Docking Filtering Problem,” in Genetic Programming Theory and Practice II, U.-M. O’Reilly, T. Yu, R. Riolo, and B. Worzel, Eds. Boston, MA: Springer US, 2005, pp. 225–244. doi: 10.1007/0-387-23254-0\_14.
	\bibitem{ref34}E. Real, C. Liang, D. So, and Q. Le, “AutoML-Zero: Evolving Machine Learning Algorithms From Scratch,” in Proceedings of the 37th International Conference on Machine Learning, Jul. 2020, vol. 119, pp. 8007–8019. [Online]. Available: https://proceedings.mlr.press/v119/real20a.html
	\bibitem{ref35}L. J. Ba, J. R. Kiros, and G. E. Hinton, “Layer Normalization,” CoRR, vol. abs/1607.06450, 2016, [Online]. Available: http://arxiv.org/abs/1607.06450
	\bibitem{ref36}Y. N. Dauphin, A. Fan, M. Auli, and D. Grangier, “Language Modeling with Gated Convolutional Networks,” in Proceedings of the 34th International Conference on Machine Learning, International Convention Centre, Sydney, Australia, Aug. 2017, vol. 70, pp. 933–941. [Online]. Available: http://proceedings.mlr.press/v70/dauphin17a.html
	\bibitem{ref37}J. Pennington, R. Socher, and C. Manning, “Glove: Global Vectors for Word Representation,” in Proceedings of the 2014 Conference on Empirical Methods in Natural Language Processing (EMNLP), Doha, Qatar, 2014, pp. 1532–1543. doi: 10.3115/v1/D14-1162.
	\bibitem{ref38}R. Socher et al., “Recursive Deep Models for Semantic Compositionality Over a Sentiment Treebank,” in Proceedings of the 2013 Conference on Empirical Methods in Natural Language Processing, Seattle, Washington, USA, 2013, pp. 1631–1642. [Online]. Available: https://www.aclweb.org/anthology/D13-1170
	\bibitem{ref39}B. Pang and L. Lee, “Seeing stars: exploiting class relationships for sentiment categorization with respect to rating scales,” in Proceedings of the 43rd Annual Meeting on Association for Computational Linguistics  - ACL ’05, Ann Arbor, Michigan, 2005, pp. 115–124. doi: 10.3115/1219840.1219855.
	\bibitem{ref40}A. L. Maas, R. E. Daly, P. T. Pham, D. Huang, A. Y. Ng, and C. Potts, “Learning Word Vectors for Sentiment Analysis,” in Proceedings of the 49th Annual Meeting of the Association for Computational Linguistics: Human Language Technologies - Volume 1, USA, 2011, pp. 142–150.
	\bibitem{ref41}X. Zhang, J. Zhao, and Y. LeCun, “Character-level Convolutional Networks for Text Classification,” in Advances in Neural Information Processing Systems, 2015, vol. 28, pp. 649–657. [Online]. Available: https://proceedings.neurips.cc/paper/2015/file/250cf8b51c773f3f8dc8b4be867a9a02-Paper.pdf
	\bibitem{ref42}T. Wolf et al., “Transformers: State-of-the-Art Natural Language Processing,” in Proceedings of the 2020 Conference on Empirical Methods in Natural Language Processing: System Demonstrations, Online, 2020, pp. 38–45. doi: 10.18653/v1/2020.emnlp-demos.6.
\end{thebibliography}
\end{document}